\pdfoutput=1

\documentclass[11pt]{article}

\usepackage[preprint]{acl}

\usepackage{times}
\usepackage{latexsym}
\usepackage{amsmath}
\usepackage{amsthm,amsmath,amssymb}
\usepackage{mathrsfs}
\usepackage{graphicx}
\usepackage{tabularx}
\usepackage{color}
\usepackage{booktabs}
\usepackage{tablefootnote}
\usepackage{pifont} 
\usepackage{arydshln} 
\usepackage{multirow}
\usepackage{hhline}
\usepackage[T1]{fontenc}

\usepackage[utf8]{inputenc}

\usepackage{microtype}

\usepackage{inconsolata}
\usepackage{tcolorbox}
\definecolor{hidden-draw}{RGB}{20,68,106}
\definecolor{hidden-pink}{RGB}{255,245,247}
\definecolor{maroon}{RGB}{148,78,99}
\definecolor{hidden-white}{RGB}{245,238,230}
\definecolor{hidden-yellow}{RGB}{255,248,227}
\definecolor{hidden-orange}{RGB}{243,215,202}
\definecolor{xm-purple}{RGB}{216, 218, 237}
\definecolor{xm-grey}{RGB}{242,242,242}
\newtcolorbox[list inside=prompt,auto counter]{prompt}[1][]{
    colbacktitle=xm-purple!90,
    colback =xm-grey!30,
    coltitle=black,
    fontupper=\footnotesize,
    boxsep=5pt,
    left=0pt,
    right=0pt,
    top=0pt,
    bottom=0pt,
    boxrule=0.5pt,
    #1,
}
\title{CoCo-Agent: A Comprehensive Cognitive MLLM Agent \\ for Smartphone GUI Automation}

\author{Xinbei Ma$^{1,2,3,4}$, Zhuosheng Zhang$^{1,*}$, Hai Zhao$^{1,2,3,4}$\thanks{$\ $Corresponding authors. This research was supported by the Joint Research Project of
Yangtze River Delta Science and Technology Innovation Community (No.
2022CSJGG1400), the Joint Funds of the National Natural Science Foundation of China (Grant No. U21B2020).} \\
$^1$School of Electronic Information and Electrical Engineering, Shanghai Jiao Tong University\\
  $^2$Department of Computer Science and Engineering, Shanghai Jiao Tong University
  \\ $^3$Key Laboratory of Shanghai Education Commission for Intelligent Interaction \\
and Cognitive Engineering, Shanghai Jiao Tong University \\ $^4$Shanghai Key Laboratory of Trusted Data Circulation and Governance in Web3 \\
\texttt{sjtumaxb@sjtu.edu.cn, zhangzs@sjtu.edu.cn, zhaohai@cs.sjtu.edu.cn}\\
}

\begin{document}
\maketitle
\begin{abstract}
Multimodal large language models (MLLMs) have shown remarkable potential as human-like autonomous language agents to interact with real-world environments, especially for graphical user interface (GUI) automation.
However, those GUI agents require comprehensive cognition ability including exhaustive perception and reliable action response.
We propose a \underline{Co}mprehensive \underline{Co}gnitive LLM \underline{Agent}, CoCo-Agent, with two novel approaches, comprehensive environment perception (CEP) and conditional action prediction (CAP), to systematically improve the GUI automation performance. 
First, CEP facilitates the GUI perception through different aspects and granularity, including screenshots and complementary detailed layouts for the visual channel and historical actions for the textual channel.
Second, CAP decomposes the action prediction into sub-problems: action type prediction and action target conditioned on the action type.
With our technical design, our agent achieves new state-of-the-art performance on AITW and META-GUI benchmarks, showing promising abilities in realistic scenarios. Code is available at \url{https://github.com/xbmxb/CoCo-Agent}.

\end{abstract}

\section{Introduction}

Graphical user interface (GUI) automation aims to enable human-like operations on operating systems with artificial intelligence instead of human efforts.
Large language models (LLMs) have demonstrated commendable performance as human-like agents, showing emergent abilities of perceiving \cite{yao2023react}, reasoning \cite{li2023camel, Park2023GenerativeAgents}, and acting \cite{wang2023voyager, autogpt}.
With the multimodal enhancement, MLLM agents become promising autonomous GUI assistants to deal with complex tasks on behalf of human operators.
To interact with the GUI environment, those agents require comprehensive cognition, including exhaustive perception and reliable action response.

Current vital challenges for autonomous agents lie in two aspects. One is \textbf{(i) the dependence on strong (M)LLMs}, and the other is \textbf{(ii) the insufficient GUI environment modeling}.

Although \textit{strong (M)LLMs} like GPT-4V \cite{openai2023gpt4} and ChatGPT \cite{ouyang2022training} ignite the development of autonomous agents, they exhibit shortcomings in realistic use.
First, the alignment requires a trustworthy domain transfer as there is a large disparity between GUI commands and natural languages. 
GUI agents are expected to generate accurate and well-formulated responses as executable GUI commands, which is non-trivial for zero-shot prompting.
For example, given the GUI that accepts commands as ``\texttt{\{action: click, touch\_point:[y$_0$, x$_0$], touch\_point:[y$_1$, x$_1$], typed\_text: `'\}}'', semantically equivalent generations like ``\texttt{Open the address book on your phone}'' is plausible but unavailable.
Second, the black-box APIs are likely to cause unexpected safety issues.
Risks to privacy and integrity may arise when granting personal device authority to a black-box API.
This significantly reduces realistic usability.
Third, the performance mainly relies on the prompt design.
The issues mentioned above leave heavy burdens on the design of prompt lines for those agents.
Besides necessary environment descriptions, the prompts (and post-processing) need to be sophisticated to enhance domain alignment, instruction following, and security risk mitigation in different circumstances.

Second, GUI agents necessitate a comprehensive multimodal perception for the modeling of the informative environment.
Existing methods for visual language models are mainly endowed with favorable abilities in semantic alignment between the vision and language modalities \cite{instructblip, ye2023mplugowl, zhao2023mmicl}. 
However, GUI contains fine-grained details and intricate semantic connections, presenting a challenge for agents to comprehend \cite{rawles2023android, li2023otterhd}. 
Consider a screenshot that includes a magnifier icon, where the conventionally accepted meaning of ``\textit{search}'' is conveyed. It implies a potential action through implicit semantics despite its small pixel size.
Thus, only leveraging general image semantics like captioning is insufficient for GUI environment modeling.
In addition, the perception of environmental information is limited by the finite input window
, where a balance between the visual and textual feature length needs to be struck. 


This work proposes CoCo-Agent, a \underline{Co}mprehensive \underline{Co}gnitive MLLM \underline{Agent}, to address the challenges above for smartphone GUI automation.
CoCo-Agent adopts a multimodal backbone of LLaVA \cite{liu2023llava} and further enhances comprehensive cognition, respectively for exhaustive perception and reliable action response.
The two proposed approaches are 
comprehensive environment perception (CEP) and conditional action prediction (CAP).
Specifically, CEP integrates GUI perception elements of textual goal, historical action, and high-level and detailed description of the vision channel.
CAP decomposes the complex and redundant GUI action commands into sub-problems following a \textit{top-down} order.
Our experiments cover diverse tasks in two GUI benchmarks, AITW \cite{rawles2023android} and META-GUI \cite{sun-etal-2022-meta}, including application manipulation, web operation, and dialogues. CoCo-Agent achieves SOTA performance with a limited parameter size.
Subsequently, we present deep analyses including element ablation, visual module selection, and future action prediction.
We show the significant effect of each perception element and the favorable choice of visual module.
We also analyze the limitations of existing datasets and illustrate the additional potential of CoCo-Agent for realistic scenarios.

Our contributions are summarized as follows:

$\circ$ We propose CoCo-Agent, an autonomous agent with comprehensive cognition for GUI, with novel approaches to enhance the perception and action response, namely comprehensive environment perception (CEP) and conditional action prediction (CAP).

$\circ$ CoCo-Agent achieves state-of-the-art performance on representative GUI benchmarks, demonstrating superior performance.

$\circ$ Extensive analyses for a systematic study of GUI automation demonstrate our significant effectiveness and realistic potential.

\section{Related Work}
This section introduces studies on autonomous language agents and multimodal perception of LLMs.

\subsection{Autonomous Language Agents}
Recent studies \cite{li2023camel, autogpt} use the term \textit{language agent} to refer to language models that interact with an environment or other agents and solve a problem.
This paper investigates the autonomous language agents that perceive the environment and then act on the environment. 

One research line relies on the strong fundamental competencies of (M)LLMs. Based on ChatGPT or GPT-4, autonomous agents can be built by only well-written prompts. Existing works have proved the reasoning, planning, and generalizing abilities of GPT-based agents, e.g., AutoGPT \cite{autogpt}, BabyAGI \cite{babyagi}, AgentGPT \cite{agentgpt}, HuggingGPT \cite{shen2023hugginggpt}, and MM-Navigator \cite{yan2023gpt}.

However, when we expect practicality and reliability, we pursue the trainable language agent that can be customized and privatized to align with given environments \cite{shao-etal-2023-character}.
Thus, another research line turns to trainable methods for open-source language models. 
m-BASH \cite{sun2022meta} adopted ROI pooling to present GUI icons in a BERT-based multi-task system.
The Auto-UI \cite{zhang2023you} was trained on a multimodal T5 \cite{2020t5}, formulating the GUI interaction to a first-principal VQA form.
CogAgent \cite{hong2023cogagent} integrated an extra attention-based high-resolution visual module with alignment pre-training.
This paper follows the second research line to discuss the trainable, open-source language agent.

\begin{figure*}[h]
    \centering
    \includegraphics[width=0.98\textwidth]{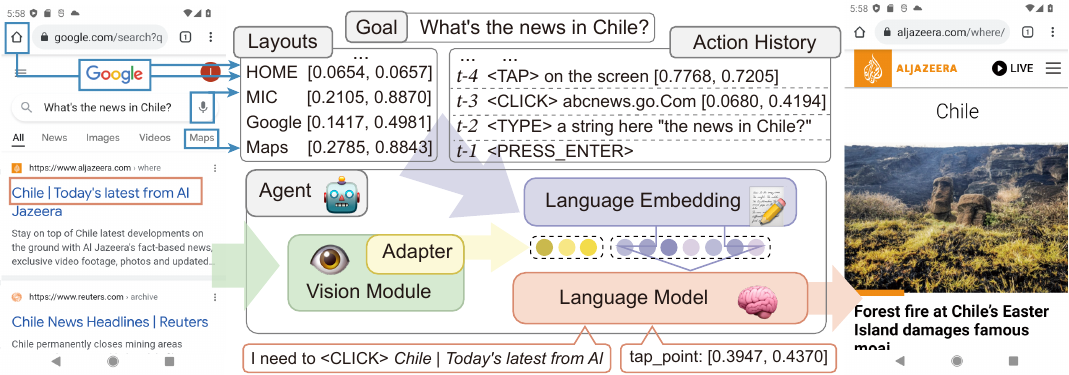}
    \caption{Overview of CoCo-Agent, illustrating the perception and action response on a time step. \textit{CEP} integrates the shown fine-grained elements. The predicted actions are formulated following \textit{CAP}. }
    \label{overview}
\end{figure*}

\subsection{Multimodal Perception}
Beyond language modeling, recent works have studied the fusion with channels of other modalities.
Because of the development of LLMs, mainstream methods usually follow a language-centric framework, i.e., encoding information of other modalities into the language embedding space. These models consist of a pre-trained encoder of other modalities, a language model, and an adapter (or a projector) as the bridge.
For example, LLaVA \cite{liu2023llava} uses a linear layer to map the vision encoding from CLIP, while BLIP-2 \cite{li2023blip} adopts a Q-former to learn a query vector to represent the image.
This endeavor has given rise to the emergence of various multimodal LLMs, such as Flamingo \cite{alayrac2022flamingo}, mPLUG \cite{ye2023mplugowl}, MiniGPT-4\&v2 \cite{zhu2023minigpt, chen2023minigptv2}, Video-LLaMA \cite{damonlpsg2023videollama}, and SpeechGPT \cite{zhang2023speechgpt}.


However, the multimodal perception is even more challenging for GUI agents.
Because GUI contains extensive detailed information with intricate semantic connections, such as very small icons conveying customary meanings (shown in Figure \ref{overview}). 
A gap remains between existing visual modules and the perception necessitated for GUI agents.

\section{Methodology}
In this section, we will first formulate the GUI automation task and then propose our CoCo-Agent. Concretely, we will describe our technical designs of cognition, namely, comprehensive environment perception (CEP) and conditional action prediction (CAP), to improve the GUI automation performance systematically. Figure \ref{overview} shows an overall illustration.
 
\subsection{Task Formalization}
The task of GUI automation is defined as an interactive sequence generation problem.
First, the user instructs the agent with a goal $g$ that can be achieved in the GUI environment in several steps.
At each step, the agent first perceives the present GUI \textit{state}, $s_t$, and predicts the next \textit{action} $a_t$, leading to the next GUI state $s_{t+1}$.
The sequential $(s, a)$ that accomplishes a goal forms an \textit{episode}. An interaction record is formulated as
\begin{equation}
\setlength{\abovedisplayskip}{5pt}
\setlength{\belowdisplayskip}{5pt}
\textsc{Record} = (g, [(s_t, a_t)]_{t=1}^n).
\label{formalization}
\end{equation}

The action space is a finite operation command set with limited parameters. 
Examples are illustrated in Table \ref{redefine}.
The state space includes any possible display from the smartphone.
As the output recipient of agents is not human but a GUI, accurate actions are expected instead of flexible expressions like natural language.

\begin{table*}[tbh]
	\centering\small
 \resizebox{\linewidth}{!}{
	{\begin{tabular}{p{4cm}p{2cm}p{1.9cm}p{1.5cm}|p{7cm}}
		\toprule
		\textbf{Action Type} & \textbf{Touch\_point}& \textbf{Lift\_point} &\textbf{Typed\_text} & \textbf{Redefined Actions in CAP} \\
            \midrule
            \midrule
             PRESS\_HOME & "[-1.0, -1.0]" & "[-1.0, -1.0]" & "" & I need to <PRESS\_HOME> \\
             PRESS\_BACK &"[-1.0, -1.0]"&"[-1.0, -1.0]"&"" & I need to <PRESS\_BACK> \\
             PRESS\_ENTER &"[-1.0, -1.0]"&"[-1.0, -1.0]"&"" & I need to <PRESS\_ENTER> \\
             STATUS\_TASK\_COMPLETE &"[-1.0, -1.0]"&"[-1.0, -1.0]"&"" & For this goal, no more action is needed, so <STATUS\_TASK\_COMPLETE> \\
             TYPE &"[-1.0, -1.0]"&"[-1.0, -1.0]"&"\{string\}" & I need to <TYPE> a string here, "typed\_text": "\{string\}" \\
             DUAL\_POINT &"\{coordinate\}"&"\{coordinate\}"&"\{string\}" & I need to <SCROLL> \{direction\} \\
             DUAL\_POINT &"\{coordinate\}"&"\{coordinate\}"&"\{string\}" & I need to <CLICK> \{item name\}, the location of \{item name\} is  "tap\_point": "\{coordinate\}" \\
             DUAL\_POINT &"\{coordinate\}"&"\{coordinate\}"&"\{string\}" & I need to <TAP> on the screen, the location is "tap\_point": "\{coordinate\}" \\
		\bottomrule
	\end{tabular}
	}}
        \caption{Illustration of JSON-formatted GUI commands in AITW (left) and our definition in CAP style (right). \textit{"[-1.0, -1.0]"} follows the default value in AITW. \textit{String}, \textit{item name}, \textit{coordinate}, and \textit{direction} are required parameters.}
	\label{redefine}
 \vspace{-1.2em}
\end{table*}

\subsection{Backbone}
Our backbone follows LLaVA \cite{liu2023llava}, which uncovers the generalization of LLM to vision modality. LLaVA consists of a Llama-2-chat-7B \cite{touvron2023llama2}, a vision encoder ($\textsc{Encoder}_{image}$), CLIP \cite{Radford2021LearningTV}, and a one-layer linear projector ($\textsc{Prj}$) to bridge the image features to the space of language embedding ($\textsc{Embed}_{text}$). 
The input is denoted as $X$, including text $X_{text}$ and image $X_{image}$, The output is denoted as $Y$.
The backbone can be formulated as 
\begin{equation}
\begin{split}
& {H}_{text} = \textsc{Embed}_{text} \textup{( } {X}_{text} \circ \hat{Y}^{0:t-1} \textup{ )}, \\
& {Z}_{image} = \textsc{Encoder}_{image} \textup{( } {X}_{image} \textup{ )}, \\
& {H}_{image} = \textsc{Prj} \textup{( } {Z}_{image} \textup{ )}, \\
& {H}_{t}^{Decoder} = \textsc{Decoder} \textup{( } {H}_{image} \circ {H}_{text}^{t} ),\\
& {P}_{t} = \textsc{LM}_{head} \textup{( } {H}_{t}^{Decoder} \textup{ )},\\
& \mathcal{L} = \sum_{t} \textsc{CE} \textup{( } {P}_{t} , {Y}_{t} \textup{ )},
\end{split}
\label{llama}
\end{equation}
where $\circ$ denotes the concatenation operation. The training objective $\mathcal{L}$ is cross entropy ($\textsc{CE}$).

\subsection{Comprehensive Environment Perception}
Environment perception is a crucial prerequisite for action responses.
The environment can be simplified to only a GUI screenshot \cite{zhang2023you}, which is highly subject to the upper-bound ability of the vision encoder.
However, there is a bottleneck for the vision channel. 
First, the size of the encoder is restricted to a relatively low resolution, e.g., 224 $\times$ 224. 
Second, the existing pre-training objectives on vision encoders mainly focus on image captioning \cite{Radford2021LearningTV, li2023blip}, which is general, high-level semantic modeling. 
Thus, fine-grained information on the screen needs to be enhanced as a complement to the high-level perception.

Our proposed comprehensive environment perception fully leverages tools like optical character recognition (OCR), which gives fine-grained layouts with readable textual hints, e.g., ``\textit{ICON\_SETTINGS: [0.1783, 0.8701]}''.
Besides the global goal, $g$, the environment state is perceived from three aspects, the present screenshot, $X_{image}$, the layouts from OCR, $L$,  and the previous actions in the present episode, ${a_{t-h:t-1}}$. The total input can be noted as
\begin{equation}
\setlength{\abovedisplayskip}{5pt}
\setlength{\belowdisplayskip}{5pt}
\begin{split}
& {X}_{text} = \textsc{Prompt} \textup{( }g, L, {a_{t-h:t-1}} \textup{ )}, X_{image},
\end{split}
\label{input}
\end{equation}
where $\textsc{Prompt}$ denotes a prompt template (Appendix \ref{a1}). $h$ denotes the number of action histories involved. The layouts $L$ are listed \textit{(item name, item coordinate)}, where \textit{items} denotes OCR results.

\subsection{Conditional Action Prediction}
Regarding action response, we propose to refactor GUI actions following the thinking order.
As is shown in the left part of Table \ref{redefine}, existing GUI actions involve redundant parameters of each command, 
including the action type, the beginning coordinates, the ending coordinates, and the possible input text.
However, these parameters are not independent of each other but show significant relations. 
For example, the coordinates are based on the action type. If the action is to click on an icon, then the touch and lift coordinates are accordingly determined.
Predicting such JSON-formatted parameters would be a waste of effort.

Thus, we propose conditional action prediction.
The GUI actions are refactored for relation decomposition as illustrated in Table \ref{redefine}. 
The actions are decomposed into two sub-problems to address, (i) action type prediction and (ii) optional action target prediction conditioned on the action type prediction.
Also, we use natural language-like expressions without compromising the accuracy.
As illustrated in Table \ref{redefine}, we change the action to a prompt line \textit{step-by-step}, explicitly decomposing and clarifying those actions.
Notably, the \textit{dual\_point} action is refined into three types: (i) \textit{scroll} action, if the beginning and ending points are farther apart than the threshold \cite{rawles2023android}; (ii) \textit{click} action involving \textit{item name}, if the tap point falls in a bounding box; (iii) \textit{tap} action, if it is not a \textit{scroll} action but matches no bounding box.

In this way, the action prediction follows a \textit{top-down} order.
First, the agent decides on action types, conditioned on which the agent further decides on the target item and coordinates.  

\noindent\textbf{Normalization.}
Based on CEP and CAP, the actions are normalized to alleviate noise, which is inevitable in real-world data.
Specifically, the target coordinates of \textit{click} actions are normalized to the centroid of the bounding box from OCR. The \textit{scroll} actions are normalized into four-direction swipes \cite{zhang2023you}. 


\section{Experiments}
This section will introduce the experimental settings including the dataset, implementation details, and baselines, followed by our empirical results.

\subsection{Dataset}
The following benchmarks of GUI automation are considered in the empirical evaluation. The dataset statistics are presented in Table \ref{datastat}.

\textbf{AITW} \cite{rawles2023android} is a benchmark for smartphone GUI, containing 715K operation episodes under 30K reality intentions. Each entry includes a goal in natural language, screenshots, and actions as Eq. \ref{formalization}. Humans collect the data on various devices and operation systems in various screen resolutions.
According to the applications domain, AITW consists of five subsets: General, Install, GoogleApps, Single, and WebShopping. 

\textbf{META-GUI} \cite{sun2022meta} smartphone dataset is originally released for multimodal dialogue generation.
Differently, the agent is enabled to communicate with the user to verify the present state or further operation, e.g., ``\textit{Is this good?}''. 
Eq. \ref{formalization} is expanded with optional utterances,
\begin{equation}
\setlength{\abovedisplayskip}{5pt}
\setlength{\belowdisplayskip}{5pt}
(s_t, a_t) \rightarrow (s_t, a_t, u^{agent}_t, u^{user}_t),
\label{meta}
\end{equation}
where utterances from the agent and the user are denoted as $u^{agent}$ and $u^{user}$.
These utterances cut an episode into several dialogue turns. META-GUI consists of 1k episodes with 18k steps. The data diversity lies in 11 applications of 6 domains including weather, calendar, search, etc. 

\begin{table}[htb]
	\centering\small
	{\begin{tabular}{p{1.9cm}p{1.3cm}p{1.5cm}p{1.0cm}}
		\toprule
		\textbf{AITW} & \textbf{Episode} &\textbf{Screen} & \textbf{Goal} \\
            \midrule
            General &9,476 &85,413 &545 \\
            Install &25,760 &250,058 &688\\
            Google Apps  &625,542 &4,903,601 &306\\
            Single &26,303 &85,668 &15,366\\
            Web Shopping  &28,061 &365,253 &13,473\\
            \midrule
		\textbf{META-GUI} & \textbf{Episode} &\textbf{Dial. turn} & \textbf{Screen} \\
            \midrule
            Train  &897 & 3692 & 14,539\\
            Dev &112 & 509 & 1,875\\
		\bottomrule
	\end{tabular}
	}
        \caption{Dataset statistics.}
	\label{datastat}
\end{table}

\subsection{Implementation}
Our implementation adopts LLaVA \cite{liu2023llava} with a LLaMA-2-chat-7B and a vision encoder, CLIP.\footnote{openai/clip-vit-large-patch14.\label{clip}} The maximum input length is 2048 following the vision instruct tuning. 
For the subsets of AITW, our experiments include two setups, i.e., separate training on each subset and unified training on the whole set. Details are shown in \ref{a1}.

\noindent \textbf{Metrics.} 
Accuracy is computed at each time step of all parameters as our metric.
The refactored action is parsed to JSON format and each parameter is compared to the action label following existing work \cite{rawles2023android}.
The predicted coordinate is considered correct if it falls in the same element bounding box as the labeled one or falls within a 14\% screen distance from the labeled one.
A scroll action is correct if its main direction is correct.
The accuracy of other parameters are exact match except for \textit{typed\_text} or dialogue responses.
The typed text of AITW is correct if the label is in the predicted text.
For META-GUI, F1 is computed for input text, and BLEU is computed for response generation.
One action is regarded as correct if all the JSON fields are correctly predicted.

\begin{table*}[htb]
	\centering\small
	\begin{tabular}{p{2.8cm}p{1.1cm}p{1.2cm}p{0.7cm}p{0.8cm}p{0.8cm}p{0.8cm}p{1.3cm}p{0.8cm}p{1.1cm}}
		\toprule
		\textbf{\emph{AITW}} & \textbf{API} & \textbf{Modality} &\textbf{Unified} &\textbf{Overall} &\textbf{General} & \textbf{Install} & \textbf{GoogleApps} &\textbf{Single} &\textbf{WebShop.} \\
            \midrule
            PaLM-2  &PaLM-2 &\textit{Text} &\makebox[0.7cm][c]{\ding{51}} &39.6 & -- & -- & -- & -- & --  \\
            ChatGPT & ChatGPT & \textit{Text} &\makebox[0.7cm][c]{\ding{51}} &7.72& 5.93 &4.38 &10.47& 9.39& 8.42\\
            MM-Navigator & GPT-4V & \textit{Text+Vision} & \makebox[0.7cm][c]{\ding{51}}  &50.54  &41.66  &42.64  &49.82  &72.83  &45.73 \\
            MM-Navigator$_{\textup{w/ text}}$ & GPT-4V & \textit{Text+Vision} &\makebox[0.7cm][c]{\ding{51}} &51.92 &42.44  &49.18  &48.26  &76.34  &43.35 \\
            MM-Navigator$_{\textup{w/ history}}$ & GPT-4V & \textit{Text+Vision} &\makebox[0.7cm][c]{\ding{51}}  &52.96  &43.01  &46.14  &49.18  &78.29  &48.18\\
            \hdashline
            BC & N/A & \textit{Text+Vision} & \makebox[0.7cm][c]{\ding{55}} &68.7 & -- & -- & -- & -- & -- \\
            BC $_{\textup{w/ history}}$ & N/A & \textit{Text+Vision} & \makebox[0.7cm][c]{\ding{55}} &73.1 &63.7 &77.5 &75.7 &80.3 &68.5\\
            LLaMA-2 & N/A & \textit{Text} & \makebox[0.7cm][c]{\ding{55}} &28.40 &28.56 &35.18 &30.99 &27.35 &19.92 \\
            Auto-UI$_{\textup{separate}}$ & N/A & \textit{Text+Vision} & \makebox[0.7cm][c]{\ding{55}} &74.22 &65.94 &77.62 &76.45 &81.39 &69.72 \\
            Auto-UI$_{\textup{unified}}$  & N/A & \textit{Text+Vision} & \makebox[0.7cm][c]{\ding{51}} &74.27 &68.24 &76.89 &71.37 &84.58 &70.26 \\
            CogAgent & N/A & \textit{Text+Vision} & \makebox[0.7cm][c]{\ding{55}} &76.88 & 65.38 & 78.86 & 74.95 & \textbf{93.49} & 71.73 \\
            \hdashline
            LLaVA$_{\textup{unified}}$ & N/A & \textit{Text+Vision} & \makebox[0.7cm][c]{\ding{51}} & 70.37  &58.93  & 72.41 & 70.81 & 83.73 & 65.98 \\ 
            CoCo-Agent$_{\textup{separate}}$ & N/A & \textit{Text+Vision} & \makebox[0.7cm][c]{\ding{55}} &77.82 &69.92 &80.60  &75.76 & 88.81 & 74.02\\
            CoCo-Agent$_{\textup{unified}}$ & N/A & \textit{Text+Vision} & \makebox[0.7cm][c]{\ding{51}} &\textbf{79.05} & \textbf{70.96} &\textbf{81.46} & \textbf{76.45} & 91.41 & \textbf{75.00} \\
		\bottomrule
	\end{tabular}
        \begin{tabular}{p{1.8cm}p{2.4cm}p{1.3cm}p{1.3cm}p{1.3cm}p{1.3cm}p{1.3cm}p{1.5cm}}
		\toprule
		\textbf{\textit{AITW}} & \textbf{Model} & \textbf{Action} & \textbf{Act. type} & \textbf{CoT. type} &\textbf{Item} & \textbf{Direction} & \textbf{Input(F1)} \\
            \midrule
            \multirow{2}{*}{General} & LLaVA$_{\textup{unified}}$ &58.93& 80.08 & N/A & 56.76 & 63.31 & 93.29  \\
            & CoCo-Agent$_{\textup{unified}}$  & 70.96 & 87.49& 76.72 &68.91 & 75.80& 97.10 \\
            \hdashline
            \multirow{2}{*}{Install} & LLaVA$_{\textup{unified}}$  & 72.41 & 85.11 & N/A &72.52 & 70.20  &  94.31 \\
            & CoCo-Agent$_{\textup{unified}}$ & 81.46 & 90.82& 85.12 &81.52  &80.49  & 97.36 \\
            \hdashline
            \multirow{2}{*}{GoogleApps} & LLaVA$_{\textup{unified}}$ & 70.81 & 88.49 & N/A & 65.55 & 74.95 & 98.75 \\
            & CoCo-Agent$_{\textup{unified}}$ &75.30 & 92.10 & 79.80 & 70.03 & 82.03 &  99.03  \\
            \hdashline
            \multirow{2}{*}{Single} & LLaVA$_{\textup{unified}}$  & 83.73 &88.19 & N/A & 85.63 & 83.95 &93.83 \\
            & CoCo-Agent$_{\textup{unified}}$ & 91.41& 95.34 &92.49 & 91.84 & 92.74 & 98.15 \\
            \hdashline
            \multirow{2}{*}{WebShopping} & LLaVA$_{\textup{unified}}$  & 65.98& 85.43 & N/A & 64.81 & 68.61 & 92.60 \\
            & CoCo-Agent$_{\textup{unified}}$ & 76.10 &  89.80 &80.20 &73.88  & 78.48 & 96.96  \\
		\bottomrule
	\end{tabular}
        \caption{Results on AITW. Part 1: Action accuracy, where primary setups are labeled: the reliance on API backends (``API''), the perceptual modalities (``Modality''), and the general ability across subsets (``Unified''). Part 2: Detailed parameter accuracies comparing our unified CoCo-Agent and LLaVA baseline.}
	\label{mainaitw}
\end{table*}

\begin{table*}[htb]
	\centering\small
	\begin{tabular}{p{2cm}p{0.8cm}p{1.5cm}p{1cm}p{1.3cm}p{1cm}p{1cm}p{1.4cm}p{1.9cm}}
		\toprule
		\textbf{\emph{META-GUI}} & \textbf{API} & \textbf{Modality} & \textbf{Action} & \textbf{Act. type} &\textbf{Item} & \textbf{Direction} & \textbf{Input (F1)} & \textbf{Utter. (BLEU)} \\
            \midrule
            LayoutLM &N/A & \textit{Text} &67.76 & 82.22 & 71.98 & 94.87 & 90.56 & 50.43\\
            LayoutLMv2  &N/A & \textit{Text+Vision} &64.48 & 85.60 & 64.38 & 92.95 & 70.76 & 58.20\\
            BERT &N/A & \textit{Text} &78.42 &87.52 & 82.84 & 93.59 & 97.24 & 62.19\\
            m-BASH  &N/A & \textit{Text+Vision} &82.74 & 90.80 & 85.90 & 96.42 & 94.23 &63.11\\
            \hdashline
            LLaVA  &N/A & \textit{Text+Vision} & 76.27 &  87.47 & 77.49 & 98.18 &96.06  & 67.24 \\
            LLaVA $_{\textup{w/ history}}$  &N/A & \textit{Text+Vision} &81.08&  91.68& 81.23 &  97.62&  96.93 & 66.57 \\
            CoCo-Agent &N/A & \textit{Text+Vision} & \textbf{88.27} & \textbf{92.59}  & \textbf{91.72} & \textbf{98.39} &96.15  & 65.90 \\
		\bottomrule
	\end{tabular}
        \caption{Results on META-GUI. }
	\label{mainmetagui}
\end{table*}

\subsection{Baselines}
For AITW, our proposed approach is compared with the following baselines.

$\bullet$ \textbf{Uni-modal API-based methods.}  
\citet{rawles2023android} and \citet{zhang2023you} have evaluated 5-shot performance on 
\textbf{PaLM-2} \cite{anil2023palm} and \textbf{ChatGPT} \cite{ouyang2022training}. The images are represented by pseudo HTML codes. The action target prediction is the item name or index, without verifying the coordinate numbers.

$\bullet$ \textbf{Multimodal API-based methods.}  \textbf{MM-Navigator} \cite{yan2023gpt} is a GPT-4V-based multimodal agent, achieving few-shot SOTA.

$\bullet$ \textbf{Training-based methods.} 
(i) \textbf{Behavioural Cloning} \cite{rawles2023android} is a Transformer-based specialized agent that models goals, screens, and historical information using a BERT \cite{devlin-etal-2019-bert}. 
(ii) \textbf{LLaMA-2} is shown as a representative trainable uni-modal LLM with pseudo HTML code inputs instead of images. The results are from \citet{zhang2023you}.
(iii) \textbf{Auto-UI} \cite{zhang2023you} bases on a multimodal encoder-decoder language model with T5 and BLIP. 
(iv) \textbf{CogAgent} \cite{hong2023cogagent} is a 9B-parameter visual LLM pre-trained for specializing in GUI understanding with a novel high-resolution cross module, which tops on AITW.

For META-GUI, we present baselines following \citet{sun2022meta} including \textbf{LayoutLMs} \cite{xu2020layoutlm, xu-etal-2021-layoutlmv2}, \textbf{BERT}, and \textbf{m-BASH} \cite{sun2022meta}. All of those need training.
m-BASH achieves SOTA which is a multi-task Transformer with Faster R-CNN \cite{ren2015faster} and ROI pooling for vision modeling.



\subsection{Main Results}
Tables \ref{mainaitw} and \ref{mainmetagui} present the main experimental results. 
\textbf{Our method surpasses the baselines significantly and achieves overall state-of-the-art performance (except for Single subset).} The unified model shows consistent advances compared to the separate training, indicating that the model learns generality across various situations.

The lower part of Table \ref{mainaitw} shows the detailed performance on AITW.
\textbf{CoCo-Agent is enabled to mimic the behavior patterns on GUI, while the limitations lie in predicting target items and scroll directions.}
(i) The action type scores achieve around 90\%. This high level indicates that the agent can learn the action patterns to operate GUI. The lower CoT type scores indicate that it is harder for the agent to differentiate \textit{dual point} actions. This is reasonable as these types of action can be much more flexible than others whose effects are more definite, like \textit{press back} and \textit{press home}.
(ii) The prediction of items and directions is more difficult for agents. Especially, the item accuracy is close to the action accuracy. 
(iii) Input prediction is relatively easy, and F1 scores are up to 97\%.

On META-GUI, we can also observe a significant improvement in action accuracy (12\%).
The accuracy of the item increases by a very large margin, while the action type and swipe direction accuracy is close to the perfect score. The input and utterance predictions are relatively consistent.

\section{Analysis}
This section presents further analysis and discussions. Section \ref{51} shows the effects of comprehensive cognition elements by ablation and replacement. Section \ref{52} discusses the capability of visual modules, followed by future action prediction in Section \ref{53}. Dataset features are analyzed in Section \ref{54}, including action type categories and alignment with realistic scenarios.

\subsection{Effects of Environment Elements}
\label{51}
\subsubsection{Ablation}
Our method combines CEP and CAP to characterize the GUI environment.
We ablate each CEP element along with the refactoring method of CAP to observe their significance for the CoCo-Agent.
Results are shown in Table \ref{ablation}. The improvement of each element is significant, especially for layouts (+5.82\%) and action history (+5.63\%). 

Besides the coordinates, the layouts provide the item names like icon names and texts shown on the screen. When combined with CAP, they bridge the prediction through rationales, making predictions easier than direct coordinate grounding.
Notably, although no historical screens are provided, historical actions with complete parameters lead to better scores than historical action types.


\begin{table}[thb]
	\centering\small
        \setlength{\abovecaptionskip}{0.1cm}
        \setlength{\belowcaptionskip}{-0.4cm}
	{\begin{tabular}{p{0.3cm}p{0.3cm}p{0.3cm}p{0.6cm}p{1.6cm}p{0.8cm}p{0.8cm}}
		\toprule
		Goal & Img. &CAP & Layout &History  & \textbf{General} &\textbf{META} \\
            \midrule
            \ding{51} &\ding{51} &\ding{55}& \ding{55} & 0   & 57.81 & 73.60\\
            \ding{51} &\ding{51} & \ding{51}& \ding{55} & 0   & 58.47 &76.90\\
            \ding{51} &\ding{51} & \ding{51}& \ding{51} & 0   & 64.29 &85.55\\
            \ding{51} &\ding{51} & \ding{51}& \ding{51} & 8 act. types   & 67.80 & 86.99\\
            \ding{51} &\ding{51} & \ding{51}& \ding{51} & 8 full actions  & 69.92 & 88.27\\
		\bottomrule
	\end{tabular}
	}
        \caption{Action accuracy for ablation studies on General of AITW and META-GUI (``META''). ``Act. types'' denotes only to provide history action types, while ``full actions'' denotes to provide actions with complete parameters.}
	\label{ablation}
\end{table}

\subsubsection{Replacement}
Actions on GUI can be flexible and the dependency on the environment is complex. 
Thus, we design a related task to probe the environment modeling further. 
We choose one element from the goal, the screen, the layouts, and the action history and replace it with another from a random data point. 
With such corrupted input, the agent is trained to select the replaced element and also predict the next action.

\begin{table}[htb]
	\centering\small
        \setlength{\belowcaptionskip}{-0.2cm}
	{\begin{tabular}{p{1.2cm}p{0.5cm}p{0.5cm}p{0.5cm}p{0.7cm}p{0.5cm}p{0.5cm}}
            \toprule
            Replace & None & Goal & Img & Layout & Hist. & Avg. \\
                \midrule
                Detection  &N/A &94.60 & 93.94 & 91.02 &92.69 &93.06\\
                Action  &70.96 &63.21 & 59.69 &57.55 &58.45 &59.73\\
            \bottomrule
        \end{tabular}
        }
        \caption{Results after replacement on General subset. \textit{Detection} denotes the accuracy of the replaced element selection. \textit{Action} denotes action accuracy.}
	\label{corruption}
\end{table}

The results are consistent with intuitions. 
(i) The wrong image and goal are more obvious replacements and get higher detection accuracy, while the layouts and action history are more complex to distinguish. 
(ii) Regarding action prediction, the accuracy decreases more with wrong layouts or wrong action history.
Those are hard to memorize and require complex modeling, therefore rely more on correct inputs.
(iii) The damage caused by a wrong image is limited due to the complement from layouts.
This suggests again that layouts are important fine-grained complements for the screen image, while the image gives an overall impression and the layouts describe detailed information. 
\begin{table}[thb]
	\centering\small
        \setlength{\abovecaptionskip}{0.2cm}
        \setlength{\belowcaptionskip}{-0.3cm}
	{\begin{tabular}{p{1.1cm}p{2.2cm}p{2.2cm}p{0.4cm}}
		\toprule
		  Model & Vision Encoder & LM & Acc   \\
            \midrule
            Auto-UI & BILP-2 Encoder &FLAN-Alpaca & 65.9  \\
            MMICL\textsuperscript{\ref{pre1k}} & Q-Former & Flan-T5-xxl & 56.4 \\
            mPLUG\textsuperscript{\ref{pre1k}} & Abstractor & Vicuna-7B & 53.0\\
            CogAgent & Low \& High Res. & Vicuna-7B & 65.4 \\
            LLaVA & CLIP  & Llama-2-chat-7B & 58.9\\
            Ours & CLIP \& Layout & Llama-2-chat-7B & 71.0\\
		\bottomrule
	\end{tabular}
	}
        \caption{GUI agents with different vision encoders.}
	\label{vision}
\end{table}

\begin{table}[thb]
	\centering\small
        \setlength{\abovecaptionskip}{0.2cm}
        \setlength{\belowcaptionskip}{-0.3cm}
	{\begin{tabular}{p{1.2cm}p{1.0cm}p{1.0cm}p{1.0cm}p{1.0cm}}
		\toprule
		  Train & 1-next & 3-next & 3-next & 3-next   \\
            \midrule
            Test & 1-next & 1-next & 2-next & 3-next \\
            \midrule
            Accuracy  & 70.96 & 58.90 & 43.40 & 37.74 \\
		\bottomrule
	\end{tabular}
	}
        \caption{Future action prediction accuracy.}
	\label{future}
\end{table}

\begin{table*}[htb]
	\centering\small
        \setlength{\abovecaptionskip}{0.2cm}
        \setlength{\belowcaptionskip}{-0.2cm}
        \begin{tabular}{p{1.4cm}p{1.7cm}p{0.5cm}p{1.3cm}p{0.4cm}p{0.4cm}p{0.4cm}p{0.4cm}p{0.3cm}p{0.4cm}p{0.3cm}p{0.4cm}p{0.4cm}p{0.5cm}}
		\toprule
		\textbf{Dataset} & \textbf{Model} &\multicolumn{2}{c}{\textbf{Dual Point}} & \multicolumn{2}{c}{\textbf{Text}} & \multicolumn{2}{c}{\textbf{Press Back}} & \multicolumn{2}{c}{\textbf{Press Home}} & \multicolumn{2}{c}{\textbf{Press Enter}} & \multicolumn{2}{c}{\textbf{Complete}} \\
            \midrule
            &  & Prop. &Acc& Prop. &Acc & Prop. &Acc & Prop. &Acc & Prop. &Acc & Prop. &Acc \\
            \midrule
            \multirow{2}{*}{General} & LLaVA  & \multirow{2}{*}{86.09}&54.16|84.71 & \multirow{2}{*}{10.90} &86.09 & \multirow{2}{*}{1.17}&8.16 & \multirow{2}{*}{5.36}&77.73 &  \multirow{2}{*}{2.61} &41.55 & \multirow{2}{*}{10.67} &62.53 \\
            & CoCo-Agent  &  &67.73|91.59 & & 84.34 & &15.31 & &89.76 &&57.99 & &78.19 \\
            \hdashline
            \multirow{2}{*}{Install} & LLaVA  & \multirow{2}{*}{69.26}& 72.53|90.77 &\multirow{2}{*}{11.77}  & 92.19 & \multirow{2}{*}{1.96} &14.98 & \multirow{2}{*}{5.79} & 67.38 &\multirow{2}{*}{0.81} & 12.69  &  \multirow{2}{*}{10.38}& 67.66  \\
            & CoCo-Agent  & & 80.46|93.84 & & 93.20 & & 41.35 & & 82.19& &  68.53 & & 83.15 \\
            \hdashline
            \multirow{2}{*}{GoogleApps} & LLaVA & \multirow{2}{*}{78.42}& 71.41|93.95 & \multirow{2}{*}{1.54} & 75.73 & \multirow{2}{*}{1.34} & 11.13 &\multirow{2}{*}{6.00} & 74.14 & \multirow{2}{*}{0.07} & 18.18  & \multirow{2}{*}{12.63} & 71.52 \\
            & CoCo-Agent & &75.41|95.04 & &80.11  & & 23.93 & & 86.60 & & 39.39 & & 83.40  \\
            \hdashline
            \multirow{2}{*}{Single} & LLaVA & \multirow{2}{*}{49.28} & 79.04|88.09 &\multirow{2}{*}{14.06} &89.74 & \multirow{2}{*}{0.17} & 57.14 & \multirow{2}{*}{0.23} & 89.47 & \multirow{2}{*}{4.52} & 72.43 & \multirow{2}{*}{31.74} & 90.06  \\
            & CoCo-Agent & & 88.94|96.92 & & 93.22 & &64.29 & & 100.00 & & 75.95  & &96.73\\
            \hdashline
            \multirow{2}{*}{WebShop.} & LLaVA  & \multirow{2}{*}{72.64} & 64.78|91.55 &\multirow{2}{*}{11.96} & 86.51 & \multirow{2}{*}{0.56}&14.00 & \multirow{2}{*}{3.82}& 75.67 & \multirow{2}{*}{3.29} & 36.32 & \multirow{2}{*}{7.73}& 57.08 \\
            & CoCo-Agent & &72.52|94.09 & & 88.72 & & 28.00 & & 89.15 & & 73.90 & & 74.04 \\
		\bottomrule
	\end{tabular}
        \caption{Accuracy of different types of actions. The agents are in the unified training setting. \textit{Prop.} is type proportion in datasets. \textit{Acc} is action accuracy|type accuracy for \textit{Dual Point} and action accuracy for others. }
	\label{difftype}
\end{table*}
\subsection{Visual Capability}
\label{52}
The vision encoder and projector largely influence the visual capability of GUI agents.
We compare a range of visual LMs with various vision encoders.

The results are shown in Table \ref{vision}. Our integrated CLIP with projector encodes an image to a 256-length vector with a 4096 hidden size.\textsuperscript{\ref{clip}}
With fine-grained layouts, CoCo-Agent receives exhaustive visual information.
Auto-UI uses a BLIP-2 with pooling leading to a 1-length image vector, which is integrated into language embedding by an attention-based fusion module.
Differently, MMICL adopts Q-Former to learn a 32-length query vector, and mPLUG adopts Abstractor to learn a 64-length vector. \footnote{The first 1000 samples in \textit{General}, the most difficult subset of AITW.\label{pre1k}} 
CogAgent uses an EVA2-CLIP-E \cite{sun2023eva} as a \textit{low-resolution encoder} and an EVA2-CLIP-L \cite{sun2023eva} with cross-attention layers as a \textit{high-resolution encoder}. In cross-attention layers, the high-resolution image features interact with each layer of the language model.

The models with only an image encoder outperform those with learnable queries.
Learning queries from \textit{image-text attention} can be unsuitable for GUI tasks, as input texts are complex and different from generic captions. 
Besides the overall semantic impression of images, straightforward textual layouts can work as an even better high-resolution module for image detail enhancement.

\subsection{Future Actions}
\label{53}
This section considers a more challenging setting of $n-$next actions prediction. 
The task is much less trivial as the agent only receives the environment state $s_t$, without the perception of future states $s_{t+1:t+n-1}$.
Thus, predicting future actions $a_{t:t+n-1}$ involves harder reasoning, planning, and environment simulation.
Table \ref{future} shows the results with $n=3$. Although the next action can be predicted with 70.96\% accuracy, predicting the following actions without environmental feedback remains to be improved.

\subsection{Dataset Features} 
\label{54}
\subsubsection{Action Type Categories}
Each action type has very different proportions in AITW, which is decided by the unbalanced distribution in natural operations. For example, the \textit{click} action is the most frequent but the \textit{complete} action appears at most once in each episode.
Thus, we divide datasets into categories according to the ground truth action type. Table \ref{difftype} shows the proportion and action accuracy.
(i) \textit{Dual point} action (including click, tap, and scroll) accounts for 69.26\% - 86.09\% in long-episode tasks and accounts for around half even in the Single subset. Other types, especially \textit{press back} and \textit{press enter} consistently account for low proportions.
Such an unbalance can limit the performance of less frequent actions.
(ii) For \textit{dual point} type, the data is sufficient and type accuracy scores are up to above 90\%. The action accuracy scores are limited by the difficulties in predicting target and direction scores as shown in Table \ref{mainaitw}. 
(iii) Single subset shows better performance on all action types and more significant on less frequent actions.
This is because the samples in Single are segmented sub-goals that give clear instructions and require fewer steps.

\subsubsection{Potential for Realistic Scenarios}
\label{542}
There is a disparity between the evaluation and realistic scenarios.
The benchmarks show randomness because the actions can be stochastically chosen with different paths for the same goal.
However, when the predicted action indicates an alternative path, it is reasonable but fails to match the label.
This leads to an underestimation. Thus, the agent has extra potential in practice. 

We study the first 500 samples of General dataset to see this phenomenon. In the first 500 actions, there are 118 (23.6\%) actions whose predictions are different from the labels. Our human evaluation considers two criteria. We check the predicted action with the goal, the last action, the present screen, and the next screen to see if (i) the predicted action can result in a situation that is similar to the next screen, or (ii) the predicted action is consistent with the goal.
We observed that only 54 (10.8\%) samples strictly contradict the episode, while other 64 predictions do not betray the goal. Among the 64 samples, there are 25 (5\%) predicted actions that can lead to a similar next state. 
For example, after typing a query in the search bar of a search engine, many actions can lead to the search results including \textit{press enter}, \textit{click the magnifier icon}, or \textit{click a proper query suggestion} (Figure \ref{same} and \ref{nobetray}).

\section{Conclusion}
This paper proposes CoCo-Agent, an MLLM-based autonomous agent for smartphone GUI.
Our method facilitates comprehensive cognition with exhaustive perception and reliable action response. 
CoCo-Agent is enhanced with two approaches. Comprehensive environment perception (CEP) integrates GUI perception, and conditional action prediction (CAP) enables a decomposition of complex commands.
CoCo-Agent achieves SOTA performance in experiments on two GUI datasets.
Further analysis shows that the agent learns the behavior patterns in smartphone GUI.
The significance of the proposed enhancements is also verified.
We discuss the unbalanced category distributions and the underestimation of agent performance.
Our work reveals the promising capabilities of MLLMs as autonomous agents, especially for complex environment perception and action response.

\section*{Limitations}
We acknowledge the limitations of this work.
(i) Resource consumption. The data is of a large amount. The training process costs computational resources compared to the zero-shot methods. Whereas, the training process only needs to be conducted once. And our unified model achieves the SOTA performance. 
Efficient LLM training or inference methods can improve the balance between resource cost and performance.
(ii) More complex settings. As is shown in Section \ref{53}, the predictions of future action remain to be improved. The ultimate goal is to empower the agent to operate the full episodes in a simulated GUI environment or a realistic device.

\section*{Future Work}
The following directions for future work are proposed based on our study.
First, GUI agents require generalization ability to support new instructions, new applications, or even new operating systems. The effectiveness of CoCo-Agent in task adaptation across domains of AITW has been preliminarily validated.
Second, GUI agents require improved multimodal training strategies. Multimodal LLMs can be strengthened by integrating GUI perception into vision-language pre-training or instruction tuning.
Third, GUI agents require comprehensive measurements. The measurements can be improved to reflect different paths for the same goal in practical scenarios.
Last but not least, there is still room for performance improvement (79.05\% for AITW and 88.27\% for META-GUI) and future action reasoning and planning for full-episode prediction.

\section*{Ethics Statement}
This section presents ethics statements in the following aspects.
(i) Data Privacy. The research dataset AITW \cite{rawles2023android} has declared control over Personal Identifiable Information. The instructions contain no toxic, biased, or misleading content. The META-GUI is based on a task-oriented dialogue dataset, SMCalFlow \cite{andreas-etal-2020-task}, crowdsourced on Amazon Mechanical Turk with qualification requirements.  CoCo-Agent does not rely on LLM APIs, preserving privacy data from leakage to LLM companies.
(ii) System security. Compared to the command-line interface (CLI), the GUI is more interpretable and controllable. Since GUI actions follow human behavior, security considerations can better align with human-oriented mechanisms, which are already implemented in existing GUIs for operating systems. However, the potential risks of GUI agents have yet to be well explored \cite{yuan2024rjudge, Yang2024WatchOF}.
(iii) Potential social impacts. On the one hand, GUI automation can facilitate efficiency and save human resources for more high-level work. On the other hand, malicious actors could abuse GUI agents to achieve undesirable purposes. For practical applications of GUI agents, platforms may need to update detection, authorization, and governance mechanisms to control potential risks for social impacts.

\bibliography{custom}
\appendix
\section{Implementation Details}
\label{a1}
\textbf{Prompt Template.}
We show the prompt templates following an example. <image> is the special token for image position in LLaVA.
\begin{prompt}[title={Prompt template for GUI agent}]
    <image> 
    
    \{item name\}$_0$ location: \{item coordinate\}$_0$

    $\dots$
    
    \{item name\}$_1$ location: \{item coordinate\}$_1$

    $\dots$

    Previous Actions:  

    \{$a_{t-h}$\}

    $\dots$

    \{$a_{t-1}$\}
    
    Goal: \{$g$\}
    
    Next action:
\end{prompt}

\begin{prompt}[title={Example of input and output }]
\color{orange}{Input:}

\color{black}<image>
ICON\_HOME location: [0.0654, 0.0657]

ICON\_THREE\_DOTS location: [0.0649, 0.9213]

google.com/search?q location: [0.0689, 0.4704]

Google location: [0.1417, 0.4981]

ICON\_THREE\_BARS location: [0.1412, 0.0796]

ICON\_MIC location: [0.2105, 0.8870]

ICON\_MAGNIFYING\_GLASS location: [0.2132, 0.1074]

What's the news in Chile? location: [0.2136, 0.4648]

Al location: [0.2772, 0.0722]

News location: [0.2768, 0.2343]

Images location: [0.2789, 0.4481]

Videos location: [0.2772, 0.6759]

Maps location: [0.2785, 0.8843]

ICON\_THREE\_DOTS location: [0.3408, 0.9407]

4 location: [0.3425, 0.0667]

https://www.aljazeera.com> where location: [0.3425, 0.4120]

Chile | Today's latest from Al location: [0.3947, 0.4370]

Jazeera location: [0.4316, 0.1574]

Stay on top of Chile latest developments on location: [0.4803, 0.4667]

the ground with Al location: [0.5101, 0.2194]

Jazeera's fact-based location: [0.5079, 0.6083]

news, location: [0.5114, 0.8759]

exclusive video footage, location: [0.5395, 0.2778]

photos location: [0.5395, 0.5907]

and location: [0.5373, 0.7056]

updated... location: [0.5395, 0.8463]

ICON\_THREE\_DOTS location: [0.6132, 0.9407]

https://www.reuters.com» archive location: [0.6132, 0.3991]

Chile News Headlines |Reuters location: [0.6658, 0.4778]

Chile permanently closes location: [0.7136, 0.2889]

mining areas location: [0.7140, 0.6713]

Chile files location: [0.7421, 0.7019]

connected to giant sinkhole location: [0.7430, 0.3139]

charges against mining company for giant... location: [0.7724, 0.4694]

ICON\_THREE\_DOTS location: [0.8452, 0.9407]

https://www.independent.co.uk> topic location: [0.8461, 0.4324]

Chile location: [0.8978, 0.1167]

latest news, location: [0.8991, 0.4111]

breaking location: [0.9009, 0.7157]

- location: [0.8991, 0.2167]

stories and comMent - The location: [0.9338, 0.4241]

ICON\_NAV\_BAR\_CIRCLE location: [0.9693, 0.4963]

ICON\_NAV\_BAR\_RECT location: [0.9693, 0.7463]

ICON\_V\_BACKWARD location: [0.9697, 0.2454]

Previous Actions:I need to <PRESS\_HOME>

I need to <TAP> on the screen, the location is "tap\_point": "[0.7768, 0.7205]"

I need to <CLICK> abcnews.go.Com, the location of abcnews.go.Com on the screen is "tap\_point": "[0.0680, 0.4194]"

I need to <TYPE> a string here, "typed\_text": "Whats the news in Chile?"

I need to <TYPE> a string here, "typed\_text": ""

I need to <PRESS\_ENTER>

Goal: What's the news in Chile? Next action:

\color{orange}{Output:} 

\color{black}{I need to <CLICK> Chile | Today's latest from Al, the location of Chile | Today's latest from Al on the screen is "tap\_point": "[0.3947, 0.4370]"}

\end{prompt}
\begin{figure*}[t]
    \centering
    \includegraphics[width=0.98\textwidth]{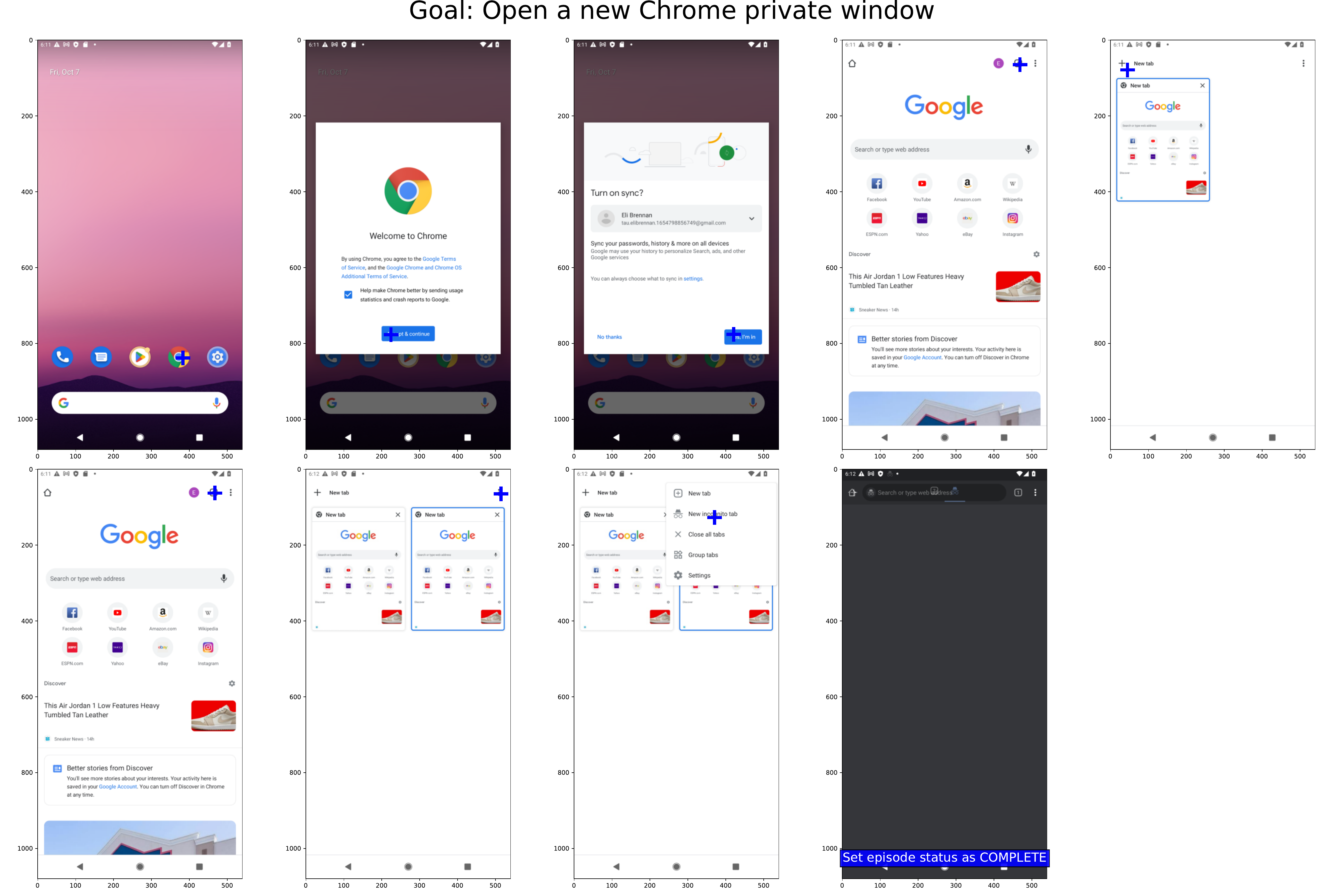}
    \caption{An example of a full episode. }
    \label{egepi}
\end{figure*}

\textbf{Experiment Details.}
For AITW, we follow the training and test set splits on \textit{General}, \textit{Install}, \textit{Google Apps}, and \textit{Web shopping} of \citet{rawles2023android}, and use the splits on \textit{Single} of \cite{zhang2023you}, which is more available.
Each subset is split for training, validation, and test sets by 8:1:1.
The first 1000 samples of the test set are used as the validation sets. The reported scores are the results of the full test sets.
For META-GUI, we follow the original dataset splits.
The maximum length of action history is 8-lastest action.
The model is trained for \{8,10,12\} epochs, with a learning rate of 2e-5. The batch size is set to \{12,16\} per device. All experiments are conducted on 4 Nvidia A800 GPUs.

\section{Examples}
\label{eg}
\begin{figure*}[h]
    \centering
    \includegraphics[width=0.98\textwidth]{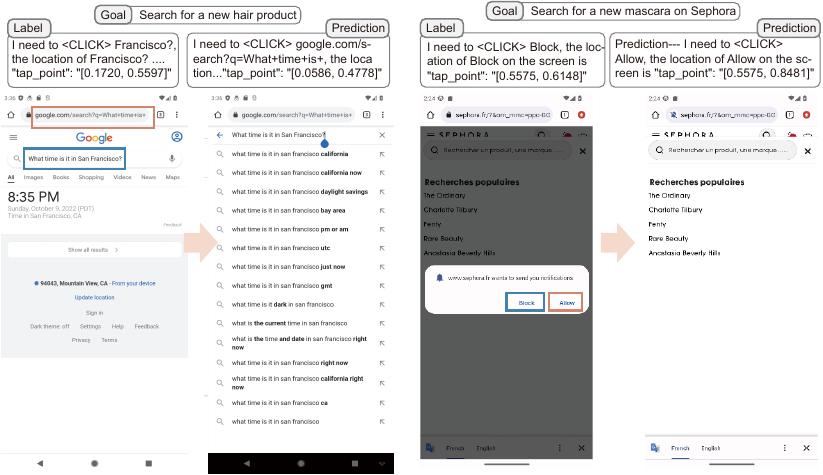}
    \caption{The predicted action leads to the same situation but does not match the label (i, in Section \ref{542}). }
    \label{same}
\end{figure*}

\begin{figure*}[h]
    \centering
    \includegraphics[width=0.98\textwidth]{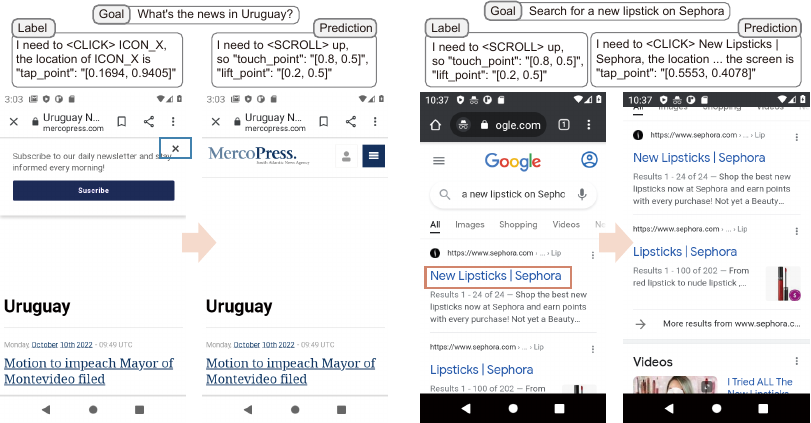}
    \caption{The predicted action is different from the label but is still reasonable for the goal (ii, in Section \ref{542}.) }
    \label{nobetray}
\end{figure*}

We show examples of a full episode in Figure \ref{egepi}.
Examples of the randomness in AITW are illustrated here. Figure \ref{same} is a case where the predicted action leads to the same situation but does not match the label (i, in Section \ref{542}).
Figure \ref{nobetray} is a case where the predicted action is different from the label but is still reasonable for the goal (ii, in Section \ref{542}). Blue boxes highlight the target items on the screen for label actions and orange boxes for predicted actions.

\section{Future Work}
The following directions for future work are proposed based on our study.
First, GUI agents require generalization ability to support new instructions, new applications, or even new operating systems. The effectiveness of CoCo-Agent in task adaptation across domains of AITW has been preliminarily validated.
Second, GUI agents require improved multimodal training strategies. Multimodal LLMs can be strengthened by integrating GUI perception into vision-language pre-training or instruction tuning.
Third, GUI agents require comprehensive measurements. The measurements can be improved to reflect different paths for the same goal in practical scenarios.

\end{document}